\documentclass[lettersize,journal]{IEEEtran}
\usepackage{amsmath,amsfonts}
\usepackage{algorithmic}
\usepackage{algorithm}
\usepackage{array}
\usepackage[caption=false,font=normalsize,labelfont=sf,textfont=sf]{subfig}
\usepackage{textcomp}
\usepackage{stfloats}
\usepackage{url}
\usepackage{verbatim}
\usepackage{graphicx}
\usepackage{cite}
\usepackage{cuted}
\usepackage{capt-of}
\usepackage{booktabs}
\usepackage{multirow}
\usepackage{amssymb}
\usepackage{hyperref}

\begin{document}

\title{
MUSON: A Reasoning-oriented Multimodal Dataset for Socially Compliant Navigation in Urban Environments
}

\author{
\IEEEauthorblockN{
    Zhuonan~Liu$^{1}$,
    Xinyu~Zhang$^{1}$,
    Zishuo~Wang$^{1}$,
    Runji~Cai$^{1}$,
    Tomohito~Kawabata$^{1}$,
    Qianyi~Li$^{1}$,
    Xuance~Peng$^{1}$,
    Tianze~Yu$^{1}$,
    Zhen~Xiong$^{1}$,
    Xuesu~Xiao$^{2}$,
    Ling Xiao$^{1,\dagger}$~\IEEEmembership{Senior Member,~IEEE}%
\thanks{$^\dagger$Corresponding author: \texttt{lingxiaodr@gmail.com} or \texttt{ling@ist.hokudai.ac.jp}.}%
\thanks{This paper is partially financially supported by JST Moonshot Goal 3 (Grant Number JPMJMS263E-12), JSPS KAKENHI (Grant No. 24K20787), and NVIDIA Academic Grant Program.}%
\thanks{$^1$Graduate School of Information Science and Technology, Hokkaido University, Sapporo, Japan.}%
\thanks{$^2$Graduate School of Computer Science, George Mason University, Fairfax County, Virginia, USA}%
}
}

\markboth{}
{Shell \MakeLowercase{}}

\maketitle

\begin{abstract}
Socially compliant navigation requires structured reasoning about dynamic pedestrians and physical constraints to ensure safe and interpretable decisions. Vision-language models (VLMs) provide a promising foundation for this task because they can integrate visual observations with language-based social knowledge. However, existing untuned VLMs still struggle to reliably understand fine-grained social norms, making task-specific fine-tuning essential. At the same time, no large-scale egocentric dataset is available this task.
To address these challenges, we introduce MUSON, a multimodal dataset for short-horizon social navigation containing 10,110 egocentric samples collected across diverse indoor and outdoor social scenes. MUSON adopts a structured five-step chain-of-thought annotation framework comprising perception, prediction, reasoning, action, and explanation. It explicitly models static physical constraints and employs a standardized six-action decision space.
Compared with existing social-navigation datasets, MUSON provides consistent annotations for reasoning, actions, and explanations. We evaluate ten representative small-to-medium VLMs on MUSON. Qwen3-VL-8B achieves the strongest decision-level performance, attaining the highest action accuracy of 0.7765 and Macro-F1 score of 0.7490, as well as the lowest collision rate of 0.0609. These results demonstrate that MUSON is an effective and reusable benchmark for advancing socially compliant navigation. The dataset is publicly available at~\url{https://github.com/MUSON-dataset/MUSON/releases/tag/v1.0}
.
\end{abstract}


\begin{figure*}[t]
\centering
\includegraphics[width=0.95\textwidth]{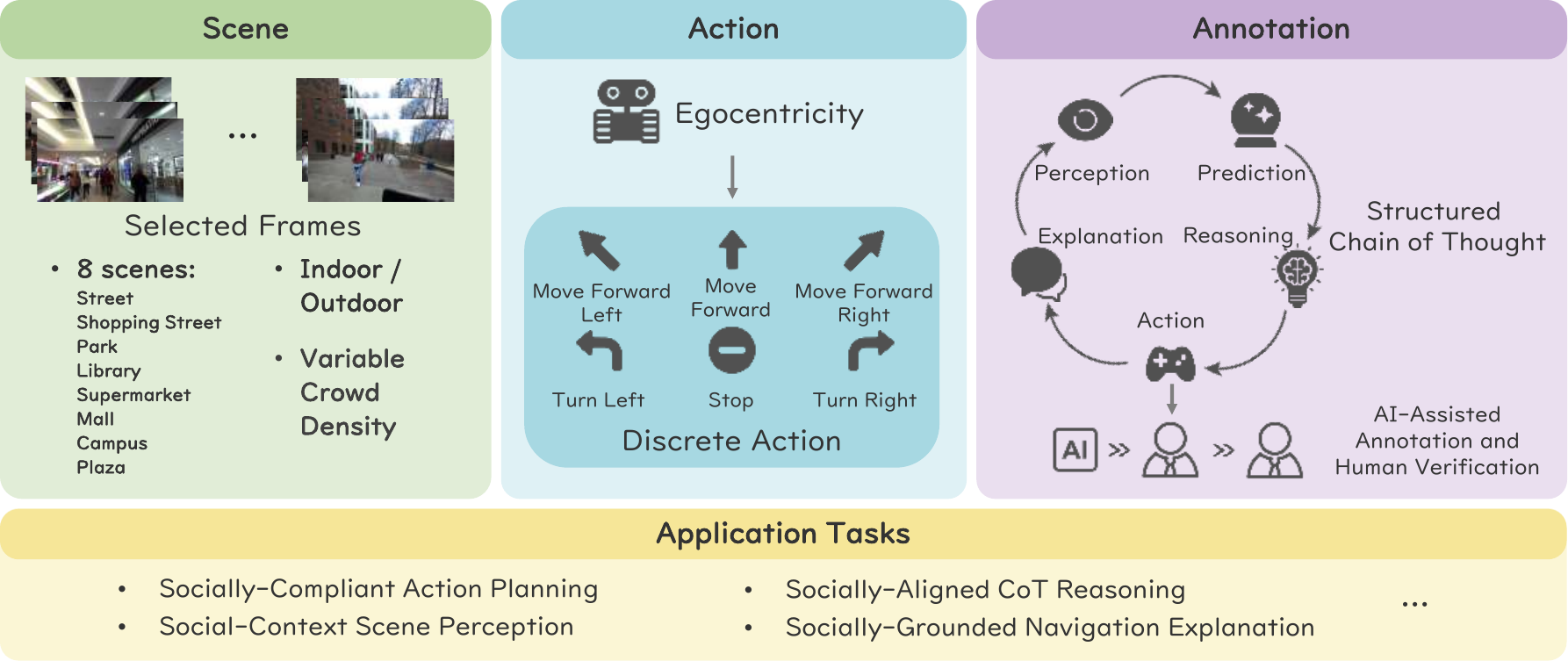}
\captionof{figure}{\textbf{The MUSON dataset construction pipeline.}
MUSON covers diverse real-world scenes across indoor and outdoor environments with varying crowd densities. Continuous navigation behaviors are discretized into six egocentric action categories with a distribution that preserves realistic forward-motion priors while retaining safety-critical corner cases. Each sample is first annotated through an AI-assisted five-step CoT pipeline and then verified, revised, and integrated by human experts to ensure annotation consistency and quality. MUSON supports multiple downstream tasks, including socially-compliant action planning, socially-aligned CoT reasoning, social-context scene perception, and socially-grounded navigation explanation.
}
\label{fig:Method}
\end{figure*}

\section{Introduction}

\IEEEPARstart{R}{esearch} in embodied navigation has traditionally focused on long-horizon, target-oriented tasks, as exemplified by benchmarks such as R2R~\cite{anderson2018vision} and RxR~\cite{ku2020room}. These benchmarks primarily address the global planning question of ``where to navigate,'' emphasizing destination reaching over local decision-making. While effective for long-range navigation, such paradigms are insufficient for scenarios that require fine-grained interaction with humans.

In contrast, socially compliant navigation emphasizes ``how to navigate'' within human-centered environments, where robots must make precise, short-horizon decisions under dynamic social constraints. This setting demands not only spatial awareness but also the ability to reason about pedestrian behaviors, shared spaces, and social norms. As a result, conventional navigation models trained on long-horizon trajectory datasets are ill-suited for social navigation tasks.

Recent efforts have begun to explore learning-based approaches for social navigation, including the use of vision language models (VLMs) to infer socially appropriate actions~\cite{sathyamoorthy2024convoi,luo2025gson,kawabata2025socialnav}. For these approaches to be effective, however, high-quality data with explicit reasoning and action supervision is essential. In particular, consistent Chain-of-Thought (CoT) annotations have recently shown promise in improving the interpretability and decision quality of VLMs to learn socially normative behaviors and ensure safety in crowded or interactive environments.

SNEI~\cite{payandeh2025social} introduces explanatory supervision for social navigation, but its limited scale and inconsistent reasoning--action alignment restrict its effectiveness as a supervised fine-tuning resource. 
Although recent socially oriented navigation datasets and benchmarks have advanced language-guided social navigation, they remain limited for training reasoning-oriented short-horizon decision models. 
SocialNav-SUB~\cite{munje2025socialnav} provides a valuable VQA-style benchmark for evaluating social scene understanding, but it is not designed to provide standardized action-level supervision for high-level navigation decisions.

To address these challenges, we propose MUSON, a large-scale reasoning-oriented dataset that jointly supports structured CoT supervision, action learning, safety evaluation, and interpretable decision-making. 
MUSON integrates high-quality annotations of both dynamic and static environmental constraints, a standardized action decision space, and diverse multi-scene data. By fine-tuning on MUSON, VLMs can acquire structured reasoning and robust action inference capabilities. Experimental results demonstrate that, when trained on MUSON, even small scale VLMs can achieve strong decision-making performance aligned with social norms.
The contributions of this work are summarized as follows: 
\begin{itemize}
    \item \textbf{A reasoning-oriented egocentric social navigation dataset.}
   We introduce MUSON, a multimodal dataset for short-horizon, socially compliant navigation, consisting of 10,110 egocentric samples collected from diverse indoor and outdoor human-shared environments. MUSON provides content-consistent five-step CoT annotations that explicitly model dynamic and static constraints and align perception, prediction, reasoning, action, and explanation within each sample.
    \item \textbf{A principled action and annotation design for social navigation.}
    MUSON discretizes continuous action space into six language-aligned action categories and adopts a standardized six-action decision space that preserves realistic navigation priors while covering rare but safety-critical interactive behaviors.
    \item \textbf{Benchmark baselines and evaluation protocol.} We benchmark ten representative Small Vision Language Models (SVLMs) on MUSON using a multidimensional protocol that evaluates decision accuracy, safety compliance, per-action robustness, and rationale similarity. We further compare action-only and structured five-step CoT supervision, showing that MUSON's reasoning annotations substantially improve short-horizon action decisions. Qwen3-VL-8B achieves the strongest decision-level performance, with an action accuracy of 0.7765, a Macro-F1 of 0.7490, and the lowest collision rate of 0.0609.
\end{itemize}

\section{Related Work}

\subsection{Dataset For Social Navigation}
Robot navigation research can be broadly categorized into long-horizon and short-horizon paradigms.
Representative works such as R2R~\cite{anderson2018vision}, RxR~\cite{ku2020room}, and VLN-CE~\cite{krantz2020beyond} have established long-horizon instruction following as the dominant setting in vision language navigation.
These datasets focus on enabling robots to understand complex natural language instructions and leverage topological memory of maps and global planning to reach distant goals~\cite{shah2023gnm, zhang2024navid}.
As a result, they often assume that low-level controllers have already solved local obstacle avoidance and physical interactions, abstracting away short-term safety and social compliance. But the actual situation is far more complex than imagined, and robots must make socially compliant policy with interactive awareness based on local and dynamic observations~\cite{martin2021jrdb}. Consequently, mastering \textit{short-horizon navigation} emerges as a fundamental prerequisite.

Earlier datasets such as SCAND~\cite{karnan2022socially} and MuSoHu~\cite{nguyen2023toward} provide real-world multimodal demonstrations and egocentric trajectories for social navigation. However, they lack structured language annotations for explicit reasoning and action supervision.
Within this setting, SNEI~\cite{payandeh2025social} represents an important effort that leverages natural language to bridge perception and socially compliant behavior.
Built upon the SCAND~\cite{karnan2022socially} dataset, SNEI introduces multi-dimensional annotations covering perception, prediction, reasoning, action, and explanation, enabling VLMs to learn human-like social reasoning in dynamic environments.
Another related benchmark is SocialNav-SUB~\cite{munje2025socialnav}, which formulates social navigation understanding as a visual question answering task to evaluate VLMs' spatial, spatiotemporal, and social reasoning abilities.
Unlike action-supervision datasets, SocialNav-SUB mainly serves as a diagnostic benchmark for scene-level understanding rather than standardized action-level supervised fine-tuning.
To support reasoning-oriented short-horizon navigation, MUSON focuses on socially compliant action decision-making with explicit five-step CoT supervision, enabling both supervised fine-tuning and safety-oriented evaluation. A detailed comparison with SNEI and SocialNav-SUB is provided in Section~\ref{subsec:comparison}.



\subsection{VLMs for Action Selection}
Translating the semantic understanding capabilities of VLMs into precise decision-making is a core challenge in embodied intelligence. Early approaches predominantly adopted end-to-end imitation learning, utilizing CNNs or recurrent neural networks (RNNs) to map raw pixels directly to actions~\cite{codevilla2018end, bansal2018chauffeurnet}. While efficient, these methods suffer from a ``black-box'' nature, lacking the causal reasoning required to generalize across unseen scenes~\cite{codevilla2019exploring}. Recently, the paradigm has shifted towards VLMs, such as PaLM-E~\cite{driess2023palm} and RT-2~\cite{zitkovich2023rt}, which leverage vast parameters for open-ended planning and reasoning~\cite{huang2022inner, yao2023react, huang2023voxposer}. Despite their strong generalization, large-scale VLMs face significant hurdles in real-world deployment: their prohibitive computational overhead and high inference latency obstruct real-time closed-loop control~\cite{vemprala2023chatgpt, wake2023chatgpt}.

To reconcile high-level reasoning with deployment efficiency~\cite{xiao2025llm}, research focus is turning toward small-scale VLMs and lightweight architectures~\cite{zhang2024tinyllama, chu2023mobilevlm, zhou2024tinyllava, lin2026moellava, bai2023qwen,xiao2026socialnav}. Simultaneously, CoT prompting has proven effective in eliciting reasoning capabilities in language models~\cite{wei2022chain}. However, effectively synergizing lightweight architectures with CoT to achieve safety-critical, reasoning-aware navigation remains an underexplored area. Moreover, existing lightweight approaches often struggle to balance the depth of reasoning with the determinacy of control. This paper investigates the potential of enabling lightweight models to master complex embodied decision-making through alignment with high-quality, structured reasoning data.

\section{The Constructed MUSON Dataset}
To address the limitations identified above, MUSON is developed with four key design considerations: expanded scale and scene coverage, high-level action-level supervision, expert-verified reasoning--action--explanation consistency, and safety-oriented interpretable evaluation. 
These considerations are instantiated through the data collection strategy, standardized action taxonomy, structured annotation schema, and evaluation tasks described in the following subsections.
\subsection{Data Collection and Scene}
\noindent\textbf{(1) Data Source and Diversity.}
Fig.~\ref{fig:Method} shows the MUSON dataset construction pipeline. MUSON is derived from the real-world first-person perspective dataset, MuSoHu~\cite{nguyen2023toward}. To capture the complexity of urban environments, we curate 10110 samples covering diverse and safety-critical scenarios. The dataset spans eight representative indoor and outdoor social environments, including streets, shopping streets, parks, libraries, supermarkets, malls, campuses, and plazas, with crowd densities ranging from sparse to dense. Table~\ref{tab:dataset_statistics} summarizes the scene and action distributions of MUSON. The scene statistics demonstrate the dataset's environmental diversity across daily public spaces, while the action statistics reflect realistic forward-motion priors and retain rare but safety-critical navigation behaviors. 
This diversity supports robust model adaptation to real-world navigation conditions.

\noindent\textbf{(2) Local Decision-Making.}
Unlike long-horizon navigation tasks that emphasize where to navigate, MUSON focuses on how to navigate within the immediate environment. By removing dependence on long-term goals or instruction history, the dataset isolates single-frame decision making and encourages models to prioritize immediate safety and social compliance in each local scene.

\noindent\textbf{(3) Ego-centric Reference Frame.}
To match the first-person nature of embodied perception, we adopt a strictly ego-centric reference frame in MUSON. All spatial descriptions are expressed using relative directions (e.g., front, front-left) anchored to the camera coordinate system, rather than absolute world coordinates. This design avoids viewpoint- and scene-specific ambiguities, improves cross-scene transferability, and ensures consistent geometric alignment between language and visual observations, facilitating robust spatial reasoning.

\subsection{Action Definition and Distribution}
\label{subsec:action}
To align with the token generation mechanism of VLMs, we discretize robot navigation behaviors into six distinct semantic actions. The action distributions are summarized in Table~\ref{tab:dataset_statistics}. 
This design converts social navigation understanding into high-level short-horizon action supervision, which is essential for supervised fine-tuning and safety-oriented decision evaluation.
This transformation into discrete action selection offers two primary advantages: 
1) Compared to continuous regression, selecting actions from a finite semantic set simplifies the prediction target, allowing the model to focus on high-level decision intentions rather than precise metric estimation.
2) Discrete actions align naturally with CoT texts, positioning the action as the logical conclusion of the reasoning process while maintaining decision consistency.

\noindent\textbf{(1) Physically Grounded Action Distribution.}
\textit{Move Forward} accounts for 54.48\% of all samples. This endows the model with necessary forward momentum and prevents overly conservative decision-making in safe scenarios.

\noindent\textbf{(2) Lateral Motion.}
\textit{Move Forward Right} and \textit{Move Forward Left} account for 22.75\% and 15.47\% of all samples, respectively. This ensures appropriate yielding and avoidance behaviors when necessary.

\noindent\textbf{(3) Preserving Safety-Critical Corner Cases.}
Although \textit{Stop} (5.49\%), \textit{Turn Right} (1.06\%), and \textit{Turn Left} (0.75\%) belong to low-frequency classes, they cover critical high-risk scenarios such as dead ends and sudden pedestrian crossings, supporting the model's ability to learn from corner cases.

\begin{table*}[t]
\centering
\caption{Dataset Statistics of MUSON, including scene and action distributions.}
\label{tab:dataset_statistics}
\resizebox{0.7\textwidth}{!}{
\begin{tabular}{l|cc|l|cc}
\hline
\multicolumn{3}{c|}{\textbf{Scene Distribution}} 
& \multicolumn{3}{c}{\textbf{Action Distribution}} \\
\hline
\textbf{Scene Category} & \textbf{Count} & \textbf{Percentage} 
& \textbf{Action Category} & \textbf{Count} & \textbf{Percentage} \\
\hline
Street          & 3669 & 36.29\% & Move Forward       & 5508 & 54.48\% \\
Mall            & 2902 & 28.70\% & Move Forward Right & 2300 & 22.75\% \\
Shopping Street & 1275 & 12.61\% & Move Forward Left  & 1564 & 15.47\% \\
Campus          & 664  & 6.57\%  & Stop               & 555  & 5.49\%  \\
Park            & 650  & 6.43\%  & Turn Right         & 107  & 1.06\%  \\
Supermarket     & 446  & 4.41\%  & Turn Left          & 76   & 0.75\%  \\
Plaza           & 379  & 3.75\%  & --                 & --   & --      \\
Library         & 125  & 1.24\%  & --                 & --   & --      \\
\hline
\textbf{Total} & \textbf{10110} & \textbf{100\%} 
& \textbf{Total} & \textbf{10110} & \textbf{100\%} \\
\hline
\end{tabular}
}
\end{table*}


\subsection{Annotation Schema}
\noindent\textbf{(1) The 5-Step CoT.} To reduce the opacity of end-to-end action prediction and strengthen reasoning--action alignment, MUSON adopts a structured five-step CoT annotation schema. Each annotation follows a complete decision chain from perception and prediction to reasoning, action selection, and explanation. This design ensures that the final action is explicitly grounded in visual evidence, anticipated pedestrian behavior, physical constraints, and social norms.

1) Perception: Fine-grained identification of dynamic entities and static constraints within the field of view that may influence navigation, including visual attributes and ego-centric spatial relationships.

2) Prediction: Forecasting future object states. Static objects are labeled as stationary, while trajectories of dynamic objects are predicted to assess potential collision risks.

3) Reasoning: A critical bridge between perception and action, where social norms and physical constraints are jointly considered to derive policy-level decisions.

4) Action: Selecting an optimal decision from a discrete action space based on the reasoning outcome.

5) Explanation: A first-person retrospective justification that summarizes the preceding chain of thought and explains the selected action.

\noindent\textbf{(2) AI-Assisted Annotation and Human Expert Verification.}
To ensure annotation consistency and scalability, we adopt an AI-assisted annotation pipeline followed by human expert verification and final integration. The pipeline consists of three stages:

1) \textit{AI-Assisted Initial Annotation.}
Structured preliminary annotations are generated for the selected frames using the GPT-5.4-mini API, following the predefined five-step schema of perception, prediction, reasoning, action, and explanation. These AI-generated annotations provide an initial structured reference for subsequent expert verification rather than serving as final ground-truth labels.

2) \textit{Human Expert Verification and Revision.}
The human expert stage is the primary annotation and verification step. Nine experts reassess the generated results based on visual evidence, annotation guidelines, and social norms. They determine the final action label as the decision judged most socially appropriate for the given scene, and correspondingly refine the perception, prediction, reasoning, and explanation components to ensure consistency across the complete annotation chain.

3) \textit{Final Consistency Check and Integration.}
After expert revision, all annotations are integrated under a unified format and subjected to a final consistency check. Samples with ambiguous actions, unsupported reasoning, inconsistent explanations, or deviations from the standardized action taxonomy are re-examined before inclusion in the final dataset. This stage ensures format consistency, reasoning--action alignment, and social-norm compliance.

\subsection{Evaluation Tasks}
Based on the structured annotation schema and diverse scenarios of MUSON, we formulate four distinct tasks to evaluate the multidimensional capabilities of navigation robots. These tasks progress from fundamental perception to complex reasoning and decision-making.

\noindent\textbf{(1) Socially-Compliant Action Planning.}
This task serves as the core control evaluation, assessing the robor's ability to make safe and normative decisions. Given an ego-centric visual observation, VLMs must predict the optimal action token $a \in \mathcal{A}$ from the six discrete semantic actions defined in Section~\ref{subsec:action}.
Unlike standard obstacle avoidance, this task emphasizes \textit{social compliance}, where the robot must respect social norms (e.g., yielding to pedestrians, maintaining distance) implicitly embedded in the dataset.


\noindent\textbf{(2) Social-Context Scene Perception.}
This task evaluates the \textit{Perception} and \textit{Prediction} components of the 5-step schema. VLMs are required to generate fine-grained text descriptions of the environment. Crucially, these descriptions must strictly adhere to the ego-centric reference frame, accurately identifying static constraints and dynamic robots using relative directional terms (e.g., ``front-left'', ``rear-right'').

\noindent\textbf{(3) Socially-Aligned CoT Reasoning.}
This task targets the \textit{Reasoning} step, acting as the bridge between perception and action. VLMs need to generate an intermediate reasoning process that weighs observed environmental factors against social norms to derive a reasoning part. This evaluates whether the model understands the \textit{why} behind a navigation scenario, rather than just fitting visual patterns to action labels.

\noindent\textbf{(4) Socially-Grounded Navigation Explanation.}
This task corresponds to the final ``Explanation'' step. VLMs need provide a retrospective natural language justification for its chosen action. Unlike the intermediate reasoning process, this explanation is user-facing, requiring the model to summarize the decision logic in a concise, human-understandable format.

\subsection{\textit{Comparison with Existing Social Navigation Datasets}}
\label{subsec:comparison}
\begin{figure*}[t]
\centering
\includegraphics[width=\textwidth]{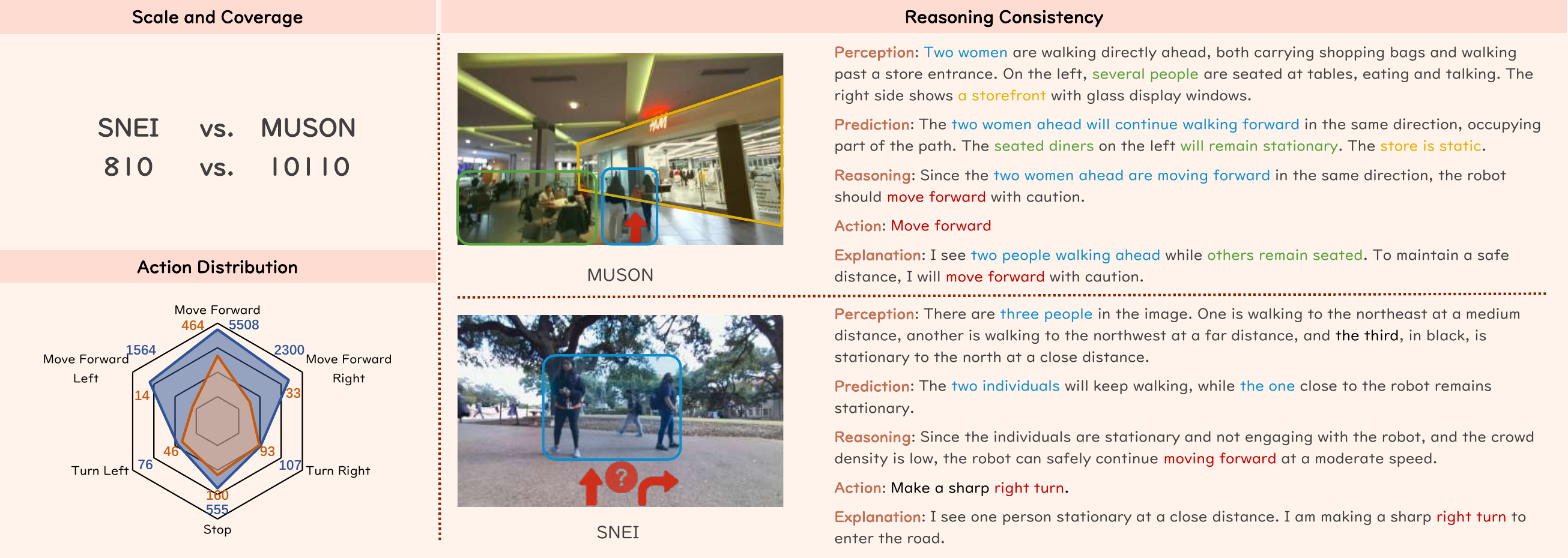}
\caption{\textbf{Visualization comparisons of samples from SNEI and MUSON.}}
\label{fig:sample}
\end{figure*}

\noindent\textbf{(1) Comparison with SNEI.}
While SNEI represents an early attempt to incorporate explanatory text into social navigation, it remains limited as a training benchmark. MUSON is designed to explicitly address these limitations (as shown in Fig.~\ref{fig:sample}):

1. Scale and Coverage. SNEI contains only 810 five-turn conversations, which is insufficient for effectively tuning vision language models. In contrast, MUSON expands the dataset to 10110 curated samples, covering a broader range of dynamic and static navigation scenarios.

2. Action Distribution. SNEI exhibits a highly long-tailed action distribution, with dominant actions overwhelming rare but safety-critical behaviors. MUSON adopts a standardized six-action decision space and improves the representation of non-forward actions, thereby enhancing coverage of diverse social interactions and safety-critical decision outcomes.

3. Reasoning Consistency. Although SNEI provides explanatory text, it lacks structured intermediate supervision, resulting in frequent inconsistencies across reasoning, actions, and explanations. MUSON enforces a structured five-step CoT annotation, ensuring logical consistency throughout perception, reasoning, action selection, and explanation.

\noindent\textbf{(2) Comparison with SocialNav-SUB.}
Beyond SNEI, SocialNav-SUB represents another related benchmark for social navigation but differs substantially from MUSON in task formulation and supervision type. SocialNav-SUB is designed as a VQA-style diagnostic benchmark for evaluating VLMs' spatial, spatiotemporal, and social reasoning abilities in navigation scenes. This formulation is useful for analyzing whether VLMs can understand human-centered navigation environments through question answering. However, SocialNav-SUB, containing only 600 samples, mainly focuses on scene-understanding evaluation rather than action-level supervision. It does not provide a standardized action decision space or structured action-level reasoning chains that explicitly connect perception, social context, and high-level navigation decisions. In contrast, MUSON provides explicit five-step CoT annotations and standardized six-action labels, making it suitable not only for evaluating social scene understanding but also for supervised training, safety-oriented evaluation, and short-horizon socially compliant action decision-making.

\section{Evaluation and Metrics}
\subsection{Safety and Social Compliance}
Following collision-related safety metrics commonly used in motion planning, we adopt collision rate (CR) to evaluate the safety of high-level navigation decisions. Unlike action accuracy, CR does not require the predicted action to be identical to the ground-truth action. Instead, it measures whether the predicted action would lead to an unsafe outcome in the given scene.
\begin{equation}
\text{CR} = \frac{1}{N} \sum_{i=1}^{N} c_i,
\end{equation}
where $N$ denotes the number of test samples and $c_i \in \{0,1\}$ is the manually assigned collision indicator for the $i$-th prediction. Specifically, $c_i=1$ if the predicted action is judged to cause a collision, and $c_i=0$ otherwise. A lower CR indicates fewer unsafe navigation decisions.

For each test sample, human evaluators inspect the visual observation and the action predicted by the model. A prediction is counted as a collision case if the selected action would cause the robot to collide with pedestrians, walls, or other physical obstacles. If the predicted action remains collision-free, it is counted as safe even when it differs from the ground-truth action.

\subsection{Concept Understanding and Reasoning Logic}
Considering decisions stem from reasoning about environmental causality, we adopt SBERTScore~\cite{reimers2019sentence} as the proxy metric for evaluating reasoning quality rather than n-gram–based metrics such as BLEU~\cite{papineni2002bleu}, which fails to capture logical and semantic consistency. Specifically, we use a pre-trained SBERT encoder to embed both predicted and reference texts into a shared high-dimensional semantic space, and compute their similarity using cosine similarity:
\begin{equation}
S_{bert} = \frac{f(T_{pred}) \cdot f(T_{gt})}{\|f(T_{pred})\| \|f(T_{gt})\|},
\end{equation}
where $T_{pred}$ denotes the explanation text generated by VLMs, $T_{gt}$ denotes the ground truth annotation provided by human experts, and $f(\cdot)$ represents the semantic embedding vector output by the SBERT encoder.

A high $S_{bert}$ indicates that the VLM generated CoT is semantically aligned with expert-provided reasoning. This suggests that the model captures meaningful physical and social cues underlying navigation decisions, rather than relying solely on surface-level action patterns.

\subsection{Decision Accuracy and Fairness}
\noindent\textbf{(1) Accuracy:} 
This metric measures the prediction correctness across all samples, which is calculated as:
\begin{equation}
Accuracy = \frac{1}{N} \sum_{i=1}^{N} \mathbb{I}(a_{pred}^i = a_{gt}^i),
\label{eg:acc}
\end{equation}
    
\noindent\textbf{(2) Macro-averaged F1 Score:} 
Given the long-tailed distribution of navigation data (dominated by move forward), the above-mentioned accuracy metric (Eq.~\ref{eg:acc}) tends to mask model failures on sparse classes. Following R2R, we employ Macro-F1 to assign equal weight to sparse actions (e.g., stop/turn left) and dominant actions:
\begin{equation}
Macro\text{-}F1 = \frac{1}{|C|} \sum_{c \in C} \frac{2 \times P_c \times R_c}{P_c + R_c},
\end{equation}
where $|C|$ is the number of action classes. $P_c$ is the precision for class $c$. $R_c$ is the recall for class $c$.

This ensures that the model's ability to handle low-frequency but high-value critical scenarios is fairly evaluated, validating its robustness under long-tailed distributions.

\noindent\textbf{(3) Per-action Accuracy:}
To further analyze model behavior on each navigation action, we report per-action accuracy for the six action classes. For action class $c$, it is calculated as:
\begin{equation}
\text{Acc}_{c} =
\frac{1}{N_c}
\sum_{i=1}^{N}
\mathbb{I}(a_{gt}^{i}=c)
\mathbb{I}(a_{pred}^{i}=a_{gt}^{i}),
\end{equation}
where $N_c$ is the number of test samples whose ground-truth action is $c$. This metric shows how reliably the model predicts each specific action, especially sparse but safety-critical actions such as \textit{stop}, \textit{turn left}, and \textit{turn right}.

\section{Experiments and Analysis}
In this section, we benchmark representative VLMs on the proposed MUSON dataset to evaluate their capabilities. We first introduce the baseline models and fine-tuning protocols, and then conduct a unified evaluation of safety, reasoning quality, and decision-making performance.

\subsection{Experimental Setup}

\noindent\textbf{(1) Base Model Selection.} To evaluate the generality and model-agnostic applicability of MUSON across different architectural paradigms and model scales, we evaluate ten representative SVLMs ranging from approximately 1B to 8B parameters. These models cover diverse architectural families and parameter scales, enabling a controlled evaluation of whether MUSON consistently benefits SVLMs with different inductive biases and capacities. Specifically, we include TinyLLaVA~\cite{zhou2024tinyllava} (Phi-2-based), NVILA~\cite{liu2025nvila}, the InternVL3 family (1B, 2B, and 8B)~\cite{zhu2025internvl3}, the Qwen2.5-VL family (3B and 7B)~\cite{bai2025qwen25vl}, and the Qwen3-VL family (2B, 4B, and 8B)~\cite{bai2025qwen3}.

\noindent\textbf{(2) Structured Prompting \& SFT Strategies.}
To overcome the issue of logical forgetting often seen in small models during complex tasks, we designed an explicit injection paradigm:

1) \textit{Prompt Engineering.} Via the System Prompt, we establish the robot's identity as a ``Safety Expert.'' We inject MUSON's three core priors: physical feasibility, social compliance, and safety first, as inviolable decision boundaries. Simultaneously, we enforce the model to follow the 5-Step CoT format, compelling it to explicitly parse environmental causality before generating action tokens.
    
\begin{figure}[t]
\centering
\includegraphics[width=0.95\linewidth]{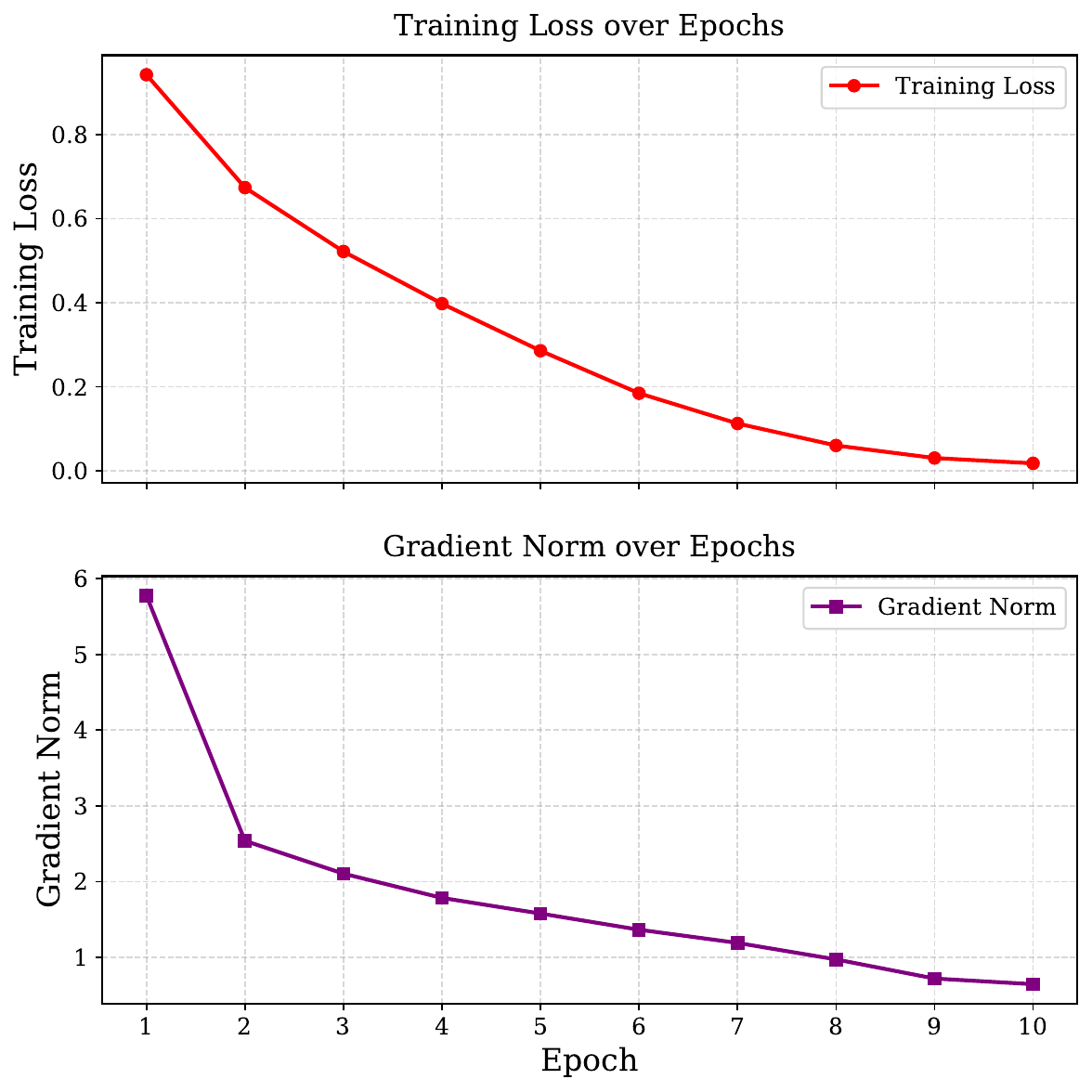}
\caption{\textbf{Training loss and gradient norm curves.} Top: Fine-tuning loss curve of Qwen3-VL-2B on MUSON. Bottom: Corresponding gradient norm curve.}
\label{fig:curve}
\end{figure}

2) \textit{Training Strategy.} We employ Full-Parameter Supervised Fine-Tuning (SFT), optimizing both the LLM and connector parameters based on a causal language modeling objective. To accelerate convergence and retain general visual features, the model is initialized from pre-trained multimodal weights. The vision tower remains frozen throughout the fine-tuning process, effectively transforming the implicit experience of human experts into explicit reasoning capabilities.As an example, Fig.~\ref{fig:curve} shows stable fine-tuning dynamics of Qwen3-VL-2B on MUSON, with steadily decreasing loss and a gradually diminishing gradient norm, indicating smooth and stable convergence.

\noindent\textbf{(3) Implementation Details.}
All experiments were conducted using the PyTorch framework with DeepSpeed optimization. Training was performed on 8 NVIDIA A100 Tensor Core GPUs. MUSON contains 10,110 annotated samples, which are split into training and testing sets with an 8:2 ratio, resulting in 8,088 training samples and 2,022 testing samples. All reported quantitative results are evaluated on the 2,022-sample test set. We employ the AdamW optimizer with a cosine learning rate schedule, setting the initial learning rate to \(2 \times 10^{-5}\) and a warmup ratio of 0.03. All input images are resized to a unified resolution of \(448 \times 448\). The per-device batch size is 4, with a gradient accumulation step of 1, resulting in a global batch size of 32 across all GPUs. Each model is fine-tuned for 10 epochs.

\subsection{Main Evaluations}
\begin{table*}[t]
\centering
\caption{\textbf{Main results on MUSON.}
We evaluate decision accuracy, safety compliance, and rationale similarity. Qwen3-VL-8B achieves the best decision performance, attaining the highest accuracy and Macro-F1 and the lowest CR. InternVL3-8B and Qwen3-VL-8B yield the strongest rationale alignment across CoT components. {\footnotesize\textit{Note:} MF, MFL, MFR, TL, TR, and Stop denote per-action accuracy for Move Forward, Move Forward Left, Move Forward Right, Turn Left, Turn Right, and Stop, respectively.}}
\label{tab:main_results}

\setlength{\tabcolsep}{3.2pt}
\renewcommand{\arraystretch}{1.08}

\resizebox{0.85\textwidth}{!}{%
\begin{tabular}{l|cc|cccccc|c}
\toprule
\multicolumn{1}{c|}{\multirow{2}{*}{\textbf{Model}}}
& \multicolumn{8}{c|}{\textbf{Decision Precision} $\uparrow$} 
& \textbf{Safety} \\
\cmidrule(lr){2-9} \cmidrule(lr){10-10}
& \textbf{Acc.} 
& \textbf{Macro-F1} 
& \textbf{MF} 
& \textbf{MFL} 
& \textbf{MFR} 
& \textbf{TL} 
& \textbf{TR} 
& \textbf{Stop} 
& \textbf{CR} $\downarrow$ \\
\midrule
TinyLLaVA (Phi-2) 
& 0.5579 & 0.2652 
& 0.7590 & 0.3012 & 0.3616 & 0.0000 & 0.0000 & 0.1500 
& 0.0885 \\

NVILA             
& 0.2681 & 0.1617 
& 0.0351 & 0.4938 & 0.6818 & 0.0000 & 0.0000 & 0.1750 
& 0.3362 \\

InternVL3-1B
& 0.5356 & 0.2526
& 0.7428 & 0.1988 & 0.3657 & 0.0000 & 0.0000 & 0.2000
& 0.1015 \\

InternVL3-2B      
& 0.5450 & 0.2792 
& 0.7077 & 0.3137 & 0.4153 & 0.0000 & 0.0625 & 0.1500 
& 0.0955 \\

InternVL3-8B
& 0.5776 & 0.2766
& 0.7725 & 0.2826 & 0.4277 & 0.0000 & 0.0000 & 0.1375
& 0.0891 \\

Qwen2.5-VL-3B     
& 0.6627 & 0.5980 
& 0.6987 & 0.5932 & 0.6343 & 0.5000 & 0.4375 & \textbf{0.6750} 
& 0.0759 \\

Qwen2.5-VL-7B     
& 0.7520 & 0.6695 
& 0.7815 & 0.6956 & \textbf{0.7438} & 0.7500 & \textbf{0.6875} & 0.6375 
& 0.0645 \\

Qwen3-VL-2B       
& 0.6899 & 0.6370 
& 0.6879 & 0.6832 & 0.7190 & \textbf{0.8750} & 0.4375 & 0.6000
& 0.0648\\

Qwen3-VL-4B       
& 0.7483 & 0.6764 
& 0.8076 & 0.6584 & 0.7273 & 0.2500 & 0.6250 & 0.4875 
& 0.0623\\

Qwen3-VL-8B       
& \textbf{0.7765} & \textbf{0.7490} 
& \textbf{0.8507} & \textbf{0.7050} & 0.6839 & 0.7500 & 0.6250 & 0.6250 
& \textbf{0.0609} \\
\bottomrule
\end{tabular}%
}
\vspace{0.2em}
\resizebox{0.65\textwidth}{!}{%
\begin{tabular}{l|cccc}
\toprule
\multicolumn{1}{c|}{\multirow{2}{*}{\textbf{Model}}}
& \multicolumn{4}{c}{\textbf{Rationale Similarity} $(S_{\mathrm{BERT}})$ $\uparrow$} \\
\cmidrule(lr){2-5}
& \textbf{Perception} 
& \textbf{Prediction} 
& \textbf{Reasoning} 
& \textbf{Explanation} \\
\midrule
TinyLLaVA (Phi-2) & 0.8042 & 0.7722 & 0.7346 & 0.7612 \\
NVILA             & 0.8124 & 0.7839 & 0.7321 & 0.7367 \\
InternVL3-1B & 0.8152 & 0.7890 & 0.7361 & 0.7651 \\
InternVL3-2B & 0.8172 & 0.7918 & 0.7392 & 0.7682 \\
InternVL3-8B & \textbf{0.8204} & 0.7993 & \textbf{0.7498} & 0.7720 \\
Qwen2.5-VL-3B     & 0.7450 & 0.5980 & 0.5360 & 0.5960 \\
Qwen2.5-VL-7B     & 0.8023 & 0.8051 & 0.7029 & 0.7657 \\
Qwen3-VL-2B       & 0.7400 & 0.7070 & 0.7080 & 0.7140 \\
Qwen3-VL-4B       & 0.8087 & 0.7469 & 0.6794 & 0.7364 \\
Qwen3-VL-8B       & 0.8191 & \textbf{0.8160} & 0.7127 & \textbf{0.7783} \\
\bottomrule
\end{tabular}%
}
\end{table*}

\begin{figure*}[htbp]
\centering
\includegraphics[width=\textwidth]{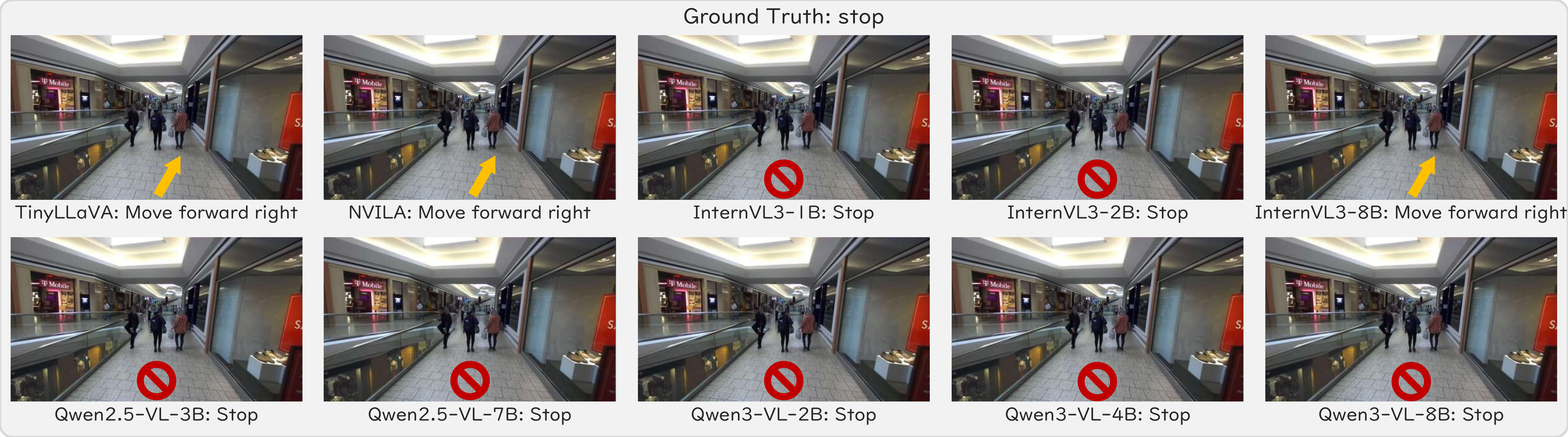}
\caption{\textbf{Visualization of high-level decision results for different SVLMs on MUSON.}}
\label{fig:action}
\vspace{0.2em}
\centering
\includegraphics[width=\textwidth]{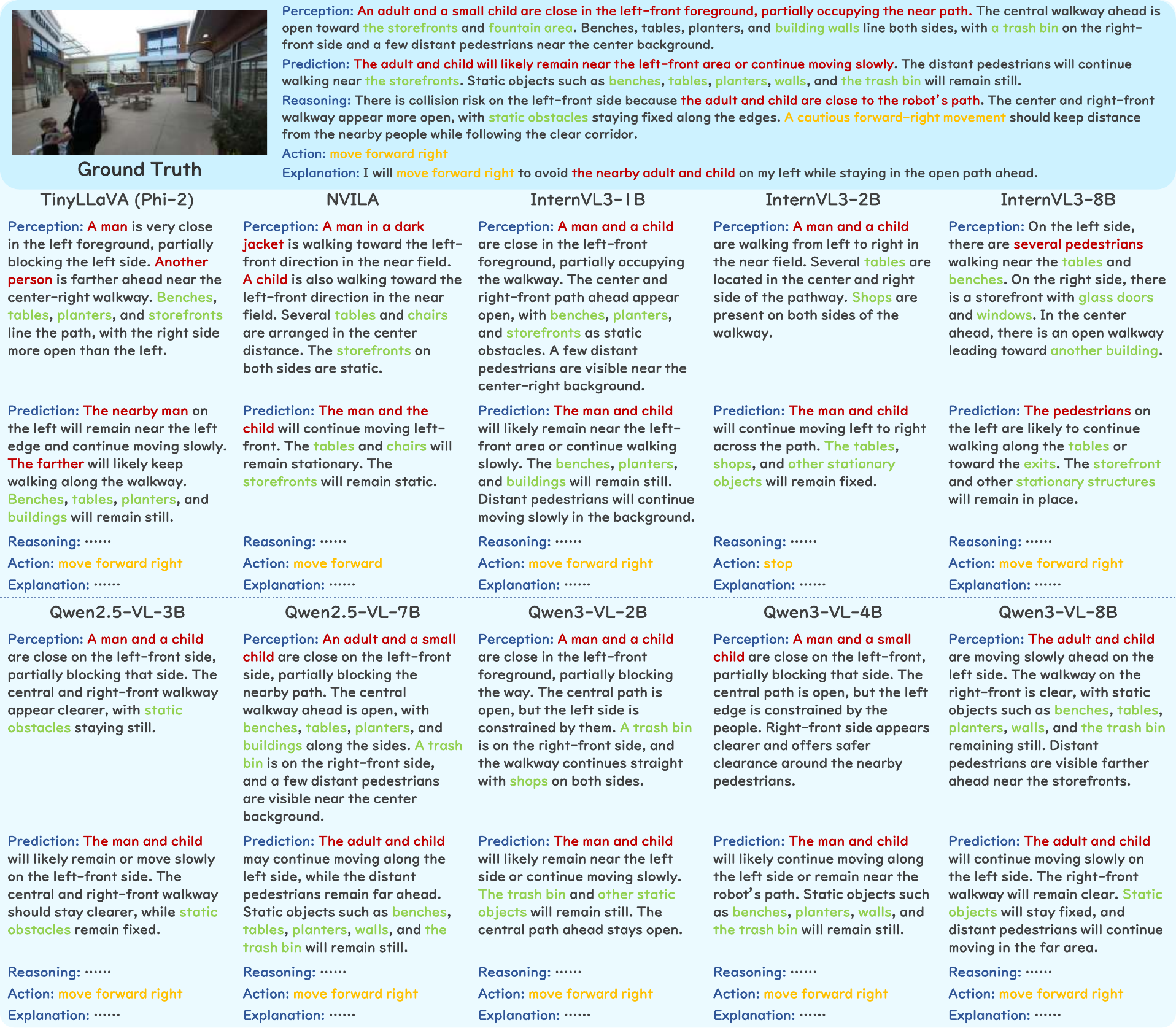}
\caption{\textbf{Visualization of high-level reasoning quality for different SVLMs on MUSON.}}
\label{fig:reasoning}
\end{figure*}

\begin{table}[t]
\centering
\caption{\textbf{Impact of CoT quality on decision accuracy.}
Qwen3-VL-2B is compared under action-only and five-turn training. Five-turn training improves MUSON accuracy but reduces SNEI performance, likely due to inconsistent reasoning, actions, and explanations in SNEI.}
\resizebox{0.8\linewidth}{!}{
\label{tab:acc_comparison}
\begin{tabular}{l c c}
\toprule
\multirow{2}{*}{\textbf{Configuration}} & \multicolumn{2}{c}{\textbf{Acc.} $\uparrow$} \\
\cmidrule(lr){2-3}
 & \textbf{SNEI} & \textbf{MUSON (Ours)} \\
\midrule
Action Only & 0.6235 & 0.5499 \\
With CoT    & 0.5741 & \textbf{0.6899} \\
\bottomrule
\end{tabular}
}
\end{table}

\noindent\textbf{(1) Comparative Analysis of SVLM Baselines.}
As shown in Table~\ref{tab:main_results}, different SVLM architectures and scales exhibit distinct navigation behaviors when trained on MUSON. Qwen3-VL-8B achieves the strongest decision-level performance, obtaining the highest accuracy and Macro-F1 as well as the lowest CR. For rationale similarity, InternVL3-8B and Qwen3-VL-8B show the strongest semantic alignment across different CoT components. NVILA shows a highly imbalanced per-action behavior, with relatively higher accuracy on \textit{Move Forward Left} and \textit{Move Forward Right} but near-zero accuracy on \textit{Move Forward}, \textit{Turn Left}, and \textit{Turn Right}. This suggests a strong action bias rather than balanced decision-making across the six-action space, which also explains its low Macro-F1 and high CR. Fig.~\ref{fig:action} presents qualitative visualizations of high-level action decisions, while Fig.~\ref{fig:reasoning} illustrates high-level semantic reasoning quality.

The per-action accuracy provides a finer-grained view of model behavior beyond overall accuracy and Macro-F1. Qwen3-VL-8B achieves the best performance on \textit{Move Forward} and \textit{Move Forward Left}, while Qwen2.5-VL-7B performs best on \textit{Move Forward Right} and \textit{Turn Right}, Qwen3-VL-2B on \textit{Turn Left}, and Qwen2.5-VL-3B on \textit{Stop}. Notably, except for the Qwen-series models, most baselines obtain near-zero accuracy on \textit{Turn Left} and \textit{Turn Right}, indicating that turning actions remain particularly difficult under long-tailed social navigation decisions. These results show that per-action accuracy is necessary to reveal class-specific weaknesses that are obscured by overall accuracy.

\noindent\textbf{(2) Reasoning Quality and Semantic Alignment.}
As shown in Table~\ref{tab:main_results}, the strongest rationale similarity is distributed across larger SVLMs. InternVL3-8B obtains the highest semantic alignment in the perception and reasoning stages, while Qwen3-VL-8B achieves the highest scores in the prediction and explanation stages. This suggests that larger models generally capture richer semantic rationales after MUSON fine-tuning, whereas Qwen3-VL-8B further translates this reasoning ability into stronger decision-level accuracy and safety compliance.

\noindent\textbf{(3) Efficacy of CoT and Dataset Validation.}
To validate the quality of MUSON and its CoT annotations, we compare Qwen3-VL-2B performance on SNEI and MUSON under \emph{Action-Only} and \emph{CoT} settings (Table~\ref{tab:acc_comparison}). The results highlight the decisive role of data quality in enabling effective reasoning for SVLMs.

On SNEI, incorporating CoT leads to an accuracy drop (0.6235 $\rightarrow$ 0.5741), indicating that its inconsistent reasoning--action alignment introduces noise rather than useful supervision. In contrast, on MUSON, CoT improves performance (0.5499 $\rightarrow$ 0.6899), demonstrating that well-structured and logically consistent annotations allow CoT to function as an effective reasoning scaffold. These results confirm that MUSON provides reliable supervision for reasoning-oriented navigation, whereas poorly aligned CoT annotations can be detrimental to capacity-limited SVLMs.

\subsection{Limitations and Future Work}

While this work represents a meaningful step forward, several limitations point to promising directions for future research.

First, rather than focusing solely on dataset scale, future work will investigate how structured annotations can more efficiently convey high-value supervisory signals under limited data budgets. In particular, we aim to refine annotation granularity and supervision design to identify which elements, such as social cues, spatial constraints, or causal rationales, most effectively enhance embodied reasoning and generalization. While the current six-action decision space is sufficient for the short-horizon social navigation setting considered in MUSON, future versions may further refine the discrete action space to better distinguish ambiguous local decisions in more complex social scenes.

Then, beyond supervised fine-tuning, we plan to explore benchmark-driven optimization of SVLMs by leveraging MUSON’s structured CoT annotations. Integrating reinforcement learning and continual adaptation mechanisms is a key direction to better align reasoning with sequential decision-making in dynamic environments, especially for lightweight models deployed under real-world constraints.

\section{Conclusion}

This paper presents MUSON, a reasoning-oriented multimodal dataset for short-horizon socially compliant navigation. MUSON contains 10,110 egocentric samples collected from diverse human-shared environments, including streets, shopping streets, parks, libraries, supermarkets, malls, campuses, and plazas. Each sample is annotated with a consistent five-step CoT schema covering perception, prediction, reasoning, action, and explanation, thereby explicitly linking visual evidence, social-context reasoning, and high-level navigation decisions. To support action-level supervision, MUSON further adopts a standardized six-action decision space that preserves realistic navigation priors while retaining rare but safety-critical behaviors.

We benchmark ten representative SVLMs on MUSON using a multidimensional evaluation protocol covering decision accuracy, safety compliance, per-action robustness, and rationale similarity. Experimental results show that Qwen3-VL-8B achieves the strongest decision-level performance, with the highest accuracy of 0.7765 and Macro-F1 of 0.7490, as well as the lowest collision rate of 0.0609. In rationale similarity evaluation, InternVL3-8B obtains the highest semantic alignment in perception and reasoning, with scores of 0.8204 and 0.7498, respectively, while Qwen3-VL-8B achieves the highest scores in prediction and explanation, with scores of 0.8160 and 0.7783, respectively. The comparison between action-only and structured CoT supervision further shows that CoT improves Qwen3-VL-2B accuracy on MUSON from 0.5499 to 0.6899, demonstrating that consistent reasoning annotations can substantially improve short-horizon socially compliant action decision-making. These results indicate that MUSON provides an effective and reusable resource for training and evaluating reasoning-oriented social navigation models.


\bibliographystyle{IEEEtran}
\bibliography{refs}

@inproceedings{anderson2018vision,
  title={Vision-and-language navigation: Interpreting visually-grounded navigation instructions in real environments},
  author={Anderson, Peter and Wu, Qi and Teney, Damien and Bruce, Jake and Johnson, Mark and S{\"u}nderhauf, Niko and Reid, Ian and Gould, Stephen and Van Den Hengel, Anton},
  booktitle={Proceedings of the IEEE Conference on Computer Vision and Pattern Recognition (CVPR)},
  pages={3674--3683},
  year={2018}
}

@inproceedings{payandeh2025social,
  title={Social-LLaVA: Enhancing Social Robot Navigation through Human-Language Reasoning},
  author={Payandeh, Amirreza and Song, Daeun and Nazeri, Mohammad and Liang, Jing and Mukherjee, Praneel and Raj, Amir Hossain and Kong, Yangzhe and Manocha, Dinesh and Xiao, Xuesu},
  booktitle={Proceedings of IEEE/RSJ International Conference on Intelligent Robots and Systems (IROS)},
  pages={17192--17198},
  year={2025}
}

@article{karnan2022socially,
  title={Socially compliant navigation dataset (scand): A large-scale dataset of demonstrations for social navigation},
  author={Karnan, Haresh and Nair, Anirudh and Xiao, Xuesu and Warnell, Garrett and Pirk, S{\"o}ren and Toshev, Alexander and Hart, Justin and Biswas, Joydeep and Stone, Peter},
  journal={IEEE Robotics and Automation Letters},
  volume={7},
  number={4},
  pages={11807--11814},
  year={2022},
}

@inproceedings{xiao2026socialnav,
  title={E-socialnav: Efficient socially compliant navigation with language models},
  author={Xiao, Ling and Song, Daeun and Xiao, Xuesu and Yamasaki, Toshihiko},
  booktitle={Proceedings of IEEE International Conference on Acoustics, Speech and Signal Processing (ICASSP)},
  pages={20077--20081},
  year={2026}
}

@inproceedings{wei2022chain,
  title={Chain-of-thought prompting elicits reasoning in large language models},
  author={Wei, Jason and Wang, Xuezhi and Schuurmans, Dale and Bosma, Maarten and Xia, Fei and Chi, Ed and Le, Quoc V and Zhou, Denny and others},
  booktitle={Proceedings of the Advances in neural information processing systems (NeurIPS)},
  volume={35},
  pages={24824--24837},
  year={2022}
}

@inproceedings{zitkovich2023rt,
  title={Rt-2: Vision-language-action models transfer web knowledge to robotic control},
  author={Zitkovich, Brianna and Yu, Tianhe and Xu, Sichun and Xu, Peng and Xiao, Ted and Xia, Fei and Wu, Jialin and Wohlhart, Paul and Welker, Stefan and Wahid, Ayzaan and others},
  booktitle={Proceedings of the Conference on Robot Learning (CoRL)},
  pages={2165--2183},
  year={2023},
}

@inproceedings{reimers2019sentence,
  title={Sentence-BERT: Sentence Embeddings using Siamese BERT-Networks},
  author={Reimers, Nils and Gurevych, Iryna},
  booktitle={Proceedings of the Conference on Empirical Methods in Natural Language Processing and the 9th International Joint Conference on Natural Language Processing (EMNLP-IJCNLP)},
  pages={3982--3992},
  year={2019}
}

@article{zhou2024tinyllava,
  title={Tinyllava: A framework of small-scale large multimodal models},
  author={Zhou, Baichuan and Hu, Ying and Weng, Xi and Jia, Junlong and Luo, Jie and Liu, Xien and Wu, Ji and Huang, Lei},
  journal={arXiv preprint arXiv:2402.14289},
  year={2024}
}

@article{bai2023qwen,
  title={Qwen technical report},
  author={Bai, Jinze and Bai, Shuai and Chu, Yunfei and Cui, Zeyu and Dang, Kai and Deng, Xiaodong and Fan, Yang and Ge, Wenbin and Han, Yu and Huang, Fei and others},
  journal={arXiv preprint arXiv:2309.16609},
  year={2023}
}

@inproceedings{driess2023palm,
  title={PaLM-E: an embodied multimodal language model},
  author={Driess, Danny and Xia, Fei and Sajjadi, Mehdi SM and Lynch, Corey and Chowdhery, Aakanksha and Ichter, Brian and Wahid, Ayzaan and Tompson, Jonathan and Vuong, Quan and Yu, Tianhe and others},
  booktitle={Proceedings of the 40th International Conference on Machine Learning (ICML)},
  pages={8469--8488},
  year={2023}
}

@article{kawabata2025socialnav,
  title={SocialNav-MoE: A Mixture-of-Experts Vision Language Model for Socially Compliant Navigation with Reinforcement Fine-Tuning},
  author={Kawabata, Tomohito and Zhang, Xinyu and Xiao, Ling},
  journal={arXiv preprint arXiv:2512.14757},
  year={2025}
}

@article{xiao2025llm,
  title={Llm-advisor: An llm benchmark for cost-efficient path planning across multiple terrains},
  author={Xiao, Ling and Yamasaki, Toshihiko},
  journal={arXiv preprint arXiv:2503.01236},
  year={2025}
}

@article{zhang2024navid,
  title={Navid: Video-based vlm plans the next step for vision-and-language navigation},
  author={Zhang, Jiazhao and Wang, Kunyu and Xu, Rongtao and Zhou, Gengze and Hong, Yicong and Fang, Xiaomeng and Wu, Qi and Zhang, Zhizheng and Wang, He},
  journal={arXiv preprint arXiv:2402.15852},
  year={2024}
}

@inproceedings{shah2023gnm,
  title={GNM: A General Navigation Model to Drive Any Robot},
  author={Shah, Dhruv and Sridhar, Ajay and Bhorkar, Arjun and Hirose, Noriaki and Levine, Sergey},
  booktitle={Proceedings of the IEEE International Conference on Robotics and Automation (ICRA)},
  pages={7226--7233},
  year={2023},
}

@inproceedings{yao2023react,
  title={REACT: SYNERGIZING REASONING AND ACTING IN LANGUAGE MODELS},
  author={Yao, Shunyu and Zhao, Jeffrey and Yu, Dian and Du, Nan and Shafran, Izhak and Narasimhan, Karthik and Cao, Yuan},
  booktitle={Proceedings of the 11th International Conference on Learning Representations (ICLR)},
  pages = {30084--30116},
  year={2023}
}

@article{martin2021jrdb,
  title={Jrdb: A dataset and benchmark of egocentric robot visual perception of humans in built environments},
  author={Martin-Martin, Roberto and Patel, Mihir and Rezatofighi, Hamid and Shenoi, Abhijeet and Gwak, JunYoung and Frankel, Eric and Sadeghian, Amir and Savarese, Silvio},
  journal={IEEE Transactions on Pattern Analysis and Machine Intelligence},
  volume={45},
  number={6},
  pages={6748--6765},
  year={2021},
}

@article{chu2023mobilevlm,
  title={Mobilevlm: A fast, strong and open vision language assistant for mobile devices},
  author={Chu, Xiangxiang and Qiao, Limeng and Lin, Xinyang and Xu, Shuang and Yang, Yang and Hu, Yiming and Wei, Fei and Zhang, Xinyu and Zhang, Bo and Wei, Xiaolin and others},
  journal={arXiv preprint arXiv:2312.16886},
  year={2023}
}

@inproceedings{huang2023voxposer,
  title={Voxposer: Composable 3d value maps for robotic manipulation with language models},
  author={Huang, Wenlong and Wang, Chen and Zhang, Ruohan and Li, Yunzhu and Wu, Jiajun and Fei-Fei, Li},
  booktitle={Proceedings of the Conference on Robot Learning (CoRL)},
  pages={540--562},
  year={2023},
}

@inproceedings{huang2022inner,
  title={Inner monologue: Embodied reasoning through planning with language models},
  author={Huang, Wenlong and Abbeel, Pieter and Pathak, Deepak and Mordatch, Igor},
  booktitle={Proceedings of the Conference on Robot Learning (CoRL)},
  pages={1769--1782},
  year={2022},
}

@inproceedings{ku2020room,
  title={Room-Across-Room: Multilingual Vision-and-Language Navigation with Dense Spatiotemporal Grounding},
  author={Ku, Alexander and Anderson, Peter and Patel, Roma and Ie, Eugene and Baldridge, Jason},
  booktitle={Proceedings of the Conference on Empirical Methods in Natural Language Processing (EMNLP)},
  pages={4379--4412},
  year={2020},
}

@inproceedings{krantz2020beyond,
  title={Beyond the nav-graph: Vision-and-language navigation in continuous environments},
  author={Krantz, Jacob and Wijmans, Erik and Majumdar, Arjun and Batra, Dhruv and Lee, Stefan},
  booktitle={Proceedings of the European Conference on Computer Vision (ECCV)},
  pages={104--120},
  year={2020}
}

@inproceedings{codevilla2018end,
  title={End-to-end driving via conditional imitation learning},
  author={Codevilla, Felipe and M{\"u}ller, Matthias and L{\'o}pez, Antonio and Koltun, Vladlen and Dosovitskiy, Alexey},
  booktitle={Proceedings of the IEEE International Conference on Robotics and Automation (ICRA)},
  pages={4693--4700},
  year={2018},
}

@article{bansal2018chauffeurnet,
  title={ChauffeurNet: Learning to drive by imitating the best and synthesizing the worst},
  author={Bansal, Mayank and Krizhevsky, Alex and Ogale, Abhijit},
  journal={arXiv preprint arXiv:1812.03079},
  year={2018}
}

@inproceedings{codevilla2019exploring,
  title={Exploring the limitations of behavior cloning for autonomous driving},
  author={Codevilla, Felipe and Santana, Eder and L{\'o}pez, Antonio M and Gaidon, Adrien},
  booktitle={Proceedings of the IEEE/CVF International Conference on Computer Vision (ICCV)},
  pages={9329--9338},
  year={2019}
}

@article{vemprala2023chatgpt,
  title={ChatGPT for robotics: Design principles and model abilities},
  author={Vemprala, Sai and Bonatti, Rogerio and Bucker, Arthur and Kapoor, Ashish},
  journal={IEEE Access},
  volume={12},
  pages={5565--5578},
  year={2024},
}

@inproceedings{wake2023chatgpt,
  title={ChatGPT empowered long-step robot control in various environments: A case application},
  author={Wake, Naoki and Kanehira, Atsushi and Sasabuchi, Kazuhiro and Takamatsu, Jun and Ikeuchi, Katsushi},
  booktitle={Proceedings of the IEEE/RSJ International Conference on Intelligent Robots and Systems (IROS)},
  pages={86--93},
  year={2023},
}

@article{zhang2024tinyllama,
  title={TinyLlama: An open-source small language model},
  author={Zhang, Peiyuan and Zeng, Guangtao and Wang, Tianduo and Lu, Wei},
  journal={arXiv preprint arXiv:2401.02385},
  year={2024}
}

@inproceedings{nguyen2023toward,
  title={Toward human-like social robot navigation: A large-scale, multi-modal, social human navigation dataset},
  author={Nguyen, Duc M and Nazeri, Mohammad and Payandeh, Amirreza and Datar, Aniket and Xiao, Xuesu},
  booktitle={Proceedings of the IEEE/RSJ International Conference on Intelligent Robots and Systems (IROS)},
  pages={7442--7447},
  year={2023},
}

@inproceedings{papineni2002bleu,
  title={Bleu: a method for automatic evaluation of machine translation},
  author={Papineni, Kishore and Roukos, Salim and Ward, Todd and Zhu, Wei-Jing},
  booktitle={Proceedings of the 40th annual meeting of the Association for Computational Linguistics (ACL)},
  pages={311--318},
  year={2002}
}

@article{luo2025gson,
  title={Gson: A group-based social navigation framework with large multimodal model},
  author={Luo, Shangyi and Sun, Peng and Zhu, Ji and Deng, Yuhong and Yu, Cunjun and Xiao, Anxing and Wang, Xueqian},
  journal={IEEE Robotics and Automation Letters},
  pages={9646--9653},
  year={2025}
}

@inproceedings{sathyamoorthy2024convoi,
  title={Convoi: Context-aware navigation using vision language models in outdoor and indoor environments},
  author={Sathyamoorthy, Adarsh Jagan and Weerakoon, Kasun and Elnoor, Mohamed and Zore, Anuj and Ichter, Brian and Xia, Fei and Tan, Jie and Yu, Wenhao and Manocha, Dinesh},
  booktitle={Proceedings of the IEEE/RSJ International Conference on Intelligent Robots and Systems (IROS)},
  pages={13837--13844},
  year={2024}
}

@inproceedings{liu2025nvila,
  title={Nvila: Efficient frontier visual language models},
  author={Liu, Zhijian and Zhu, Ligeng and Shi, Baifeng and Zhang, Zhuoyang and Lou, Yuming and Yang, Shang and Xi, Haocheng and Cao, Shiyi and Gu, Yuxian and Li, Dacheng and others},
  booktitle={Proceedings of the Computer Vision and Pattern Recognition Conference (CVPR)},
  pages={4122--4134},
  year={2025}
}

@article{bai2025qwen25vl,
  title={Qwen2.5-VL Technical Report},
  author={Bai, Shuai and Chen, Keqin and Liu, Xuejing and Wang, Jialin and Ge, Wenbin and Song, Sibo and Dang, Kai and Wang, Peng and Wang, Shijie and Tang, Jun and Zhong, Humen and Zhu, Yuanzhi and Yang, Mingkun and Li, Zhaohai and Wan, Jianqiang and Wang, Pengfei and Ding, Wei and Fu, Zheren and Xu, Yiheng and Ye, Jiabo and Zhang, Xi and Xie, Tianbao and Cheng, Zesen and Zhang, Hang and Yang, Zhibo and Xu, Haiyang and Lin, Junyang},
  journal={arXiv preprint arXiv:2502.13923},
  year={2025}
}

@article{zhu2025internvl3,
  title={Internvl3: Exploring advanced training and test-time recipes for open-source multimodal models},
  author={Zhu, Jinguo and Wang, Weiyun and Chen, Zhe and Liu, Zhaoyang and Ye, Shenglong and Gu, Lixin and Tian, Hao and Duan, Yuchen and Su, Weijie and Shao, Jie and others},
  journal={arXiv preprint arXiv:2504.10479},
  year={2025}
}

@article{bai2025qwen3,
  title={Qwen3-vl technical report},
  author={Bai, Shuai and Cai, Yuxuan and Chen, Ruizhe and Chen, Keqin and Chen, Xionghui and Cheng, Zesen and Deng, Lianghao and Ding, Wei and Gao, Chang and Ge, Chunjiang and others},
  journal={arXiv preprint arXiv:2511.21631},
  year={2025}
}

@inproceedings{munje2025socialnav,
  title={SocialNav-SUB: Benchmarking VLMs for Scene Understanding in Social Robot Navigation},
  author={Munje, Michael Joseph and Tang, Chen and Liu, Shuijing and Hu, Zichao and Zhu, Yifeng and Cui, Jiaxun and Warnell, Garrett and Biswas, Joydeep and Stone, Peter},
  booktitle={Proceedings of the Conference on Robot Learning (CoRL)},
  pages={1120--1143},
  year={2025}
}

@article{lin2026moellava,
  title={MoE-LLaVA: Mixture of experts for large vision-language models},
  author={Lin, Bin and Tang, Zhenyu and Ye, Yang and Huang, Jinfa and Zhang, Junwu and Pang, Yatian and Jin, Peng and Ning, Munan and Luo, Jiebo and Yuan, Li},
  journal={IEEE Transactions on Multimedia},
  volume={28},
  pages={4408--4419},
  year={2026}
}

\newpage

\vfill

\end{document}